# "Mind your p's and q's?":
# or the peregrinations of an apostrophe in 17<sup>th</sup> Century English

**Odile Piton**
SAMM, University of Panthéon-Sorbonne
Paris, France.
piton@univ-paris1.fr

**Hélène Pignot**
SAMM, University of Panthéon-Sorbonne
Paris, France.
helene.pignot@univ-paris1.fr

### Abstract

If the use of the apostrophe in contemporary English often marks the Saxon genitive, it may also indicate the omission of one or more letters. Some writers (wrongly?) use it to mark the plural in symbols or abbreviations, visualised thanks to the isolation of the morpheme "s". This punctuation mark was imported from the Continent in the 16<sup>th</sup> century. During the 19<sup>th</sup> century its use was standardised. However the rules of its usage still seem problematic to many, including literate speakers of English. "All too often, the apostrophe is misplaced", or "errant apostrophes are springing up everywhere" is a complaint that Internet users frequently come across when visiting grammar websites. Many of them detail its various uses and misuses, and attempt to correct the most common mistakes about it, especially its misuse in the plural, called greengrocers' apostrophes and humorously misspelled "greengrocers apostrophe's". While studying English travel accounts published in the seventeenth century, we noticed that the different uses of this symbol may accompany various models of metaplasms. We were able to highlight the linguistic variations of some lexemes, and trace the origin of modern grammar rules governing its usage.

## 1 Introduction

After 450 years of existence, the apostrophe is under attack in the English-speaking world. This epiphenomenon not only reflects the evolution of language, but also that of today's society with its heavy reliance on sophisticated technology.

Inconsistent usage of the apostrophe may indeed cause problems. The possessive apostrophe appears in addresses and on road signs, which must concord with the addresses stored in the GPS databases. The emergency services, firemen or ambulance drivers who need to find an address on GPS can no longer waste precious time wondering whether there is an apostrophe in the address of the person they are trying to rescue! As a result, in 2001 its use in public addresses was banned in Australia and in some English cities such as Birmingham where, as of 2009, St. Paul's Square is to be spelled St. Pauls Square, to the utter dismay of purists or English language pedants.[1]

The apostrophe was imported to Great Britain from the continent in the 16<sup>th</sup> century. But it was not until the 19<sup>th</sup> century that its use was standardised, according to David Crystal (1995), a distinguished linguist and the author of many books on the English language.

In the United Kingdom there is no such institution as an Academy in the French style. Therefore no strict "rules" were established regarding this matter, and they have evolved over time.[2]

Our study of a corpus of English travel accounts in the Ottoman Empire has revealed various uses of the apostrophe in the 17<sup>th</sup> century.

---

[1] See the following articles *inter alia* "Birmingham City Council bans apostrophes from road signs" (Birmingham Post, by Paul Dale Jan. 29, 2009); "Apostrophe catastrophe for city's street signs" (The Independent [UK], Jan. 30, 2009); "City drops apostrophes from signs" (BBC, Jan. 29, 2009).

2 The guidelines given by Gilbaldi (2003) pp. 90-91 are very clear and consistent. The possessive of singular proper nouns in US English is formed by adding an apostrophe, even if the noun ends in "s", as in Descartes's or Dickens's. To form the plural possessive, you only need to add an apostrophe. But US usage differs from English usage, some proper nouns such as Descartes, Socrates and Jesus only take an apostrophe and no *s* in British English. Michael Swan (1992, 505) says apostrophes may be used in the plurals of letters and numbers as in "he writes b's instead of d's" and in "it was in the early 1960's"; in contracted words ("'flu" (influenza); or in dates '79 (=1979). Gilbaldi only accepts the form "1960s".

The frequency of its use varies according to the authors and according to the dates when the texts were written. But although our texts were all written in the 17<sup>th</sup> century, we found that usage was quite coherent, as we shall see in the study presented below.

## 2 A Brief Historical Outline

In writing this brief historical outline, we are indebted to Jacques André's excellent article published in the review *Graphê* (2008), "Funeste destinée : l'apostrophe détournée" and to Lucien Febvre and Henry-Jean Martin's book, *L'Apparition du Livre* (1958).

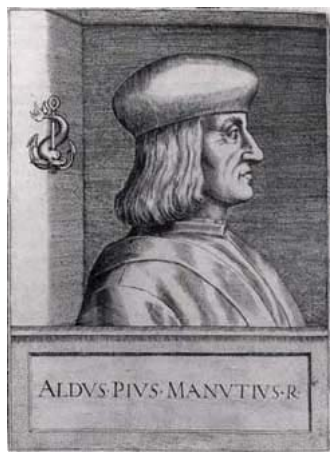

Illustration 1: Aldus Manutius, the inventor of italic type

### 2.1 From Italy to Great Britain

If the invention of movable metal printing type dates from the mid-15<sup>th</sup> century and printing quickly spread throughout Europe during the 15<sup>th</sup> century, it was not until the 16<sup>th</sup> century that the apostrophe appeared in printing type. This character, used in Greek but absent from classical Latin except in poetry, first appeared in 1501 when Aldus Manutius (Illustration 1) printed *Le cose volgari di messer Francesco Petrarcha*, the first book in Italian of his collection of *libri portatiles*, pocket-books of classical writers that he had started with an edition of Virgil published in italics. In Italian the apostrophe marked the elision of a letter.

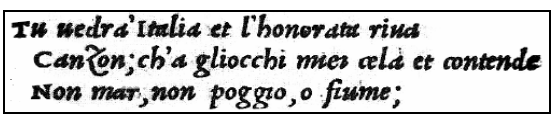

Illustration 2: Apostrophes in an Italian text

**France:** The sixteenth century saw a reform of printing in France. Printed characters were different from handwritten letters, and writing and punctuation rules were gradually set out. The printers were humanist scholars; among them Geofroy Tory, a painter, engraver and master printer of the French King François I, inspired by Italian models, published *Champ fleury* (1529) —illustration 3— a groundbreaking treatise on typography and orthography.

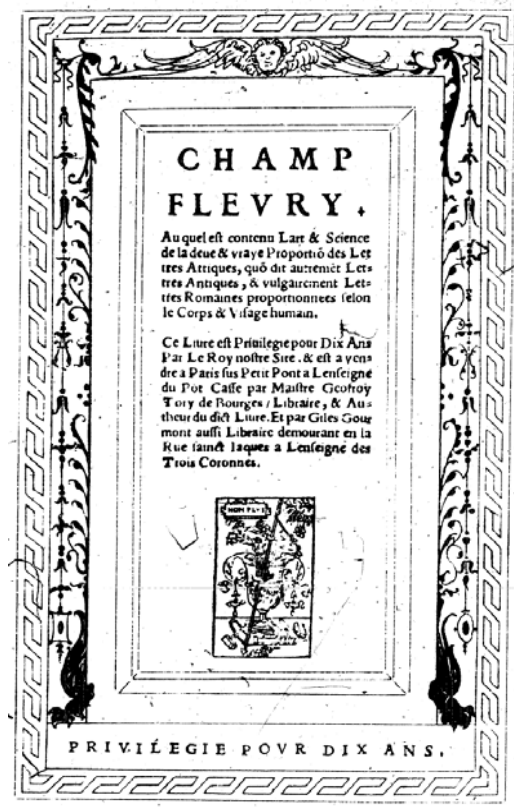

Illustration 3: G. Tory's *Champ fleury*

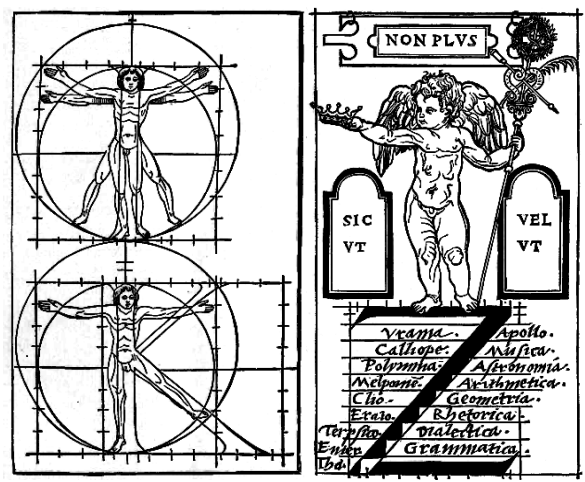

Illustration 4: G. Tory's way of forming the letters I, K & Z

In this book (illustration 4), he presents a somewhat paradoxical theory: the forms of all Roman capital letters derive from the proportions of the human body that he considers as an archetype of beauty.[3] He also proposes a reform of spelling, recommending the use of accents (at the beginning of the 16th century it was difficult to grasp the pronunciation of French as there were no accents), the cedilla and the apostrophe. In the tiers livre of *Champ fleury*, he remarks that Latin authors use a "point crochu quon appelle *apostrophus*" (a hooked dot that is called the apostrophe) to mark contraction (André 2008, 4).[4]

He applied his principles in two books *Adolescence clémentine* and *Briesve doctrine pour deuement escripre selon la propriété du langaige François* that he printed in 1533.

In 1529 the physician and grammarian Jacques Dubois published *Très utile et compendieux traicté de l'art et science dorthographie Gallicane.* In his grammar *Isagωge* (Initiation) he recommends the use of the Greek apostrophe to signal the omission of a letter. In the meantime, the apostrophe appeared in other books, its use became widespread, and its inclusion in French words obligatory: *le ami* being henceforth spelled *l'ami*.

Illustration 5: Apostrophes in a 16th century French Text

If the apostrophe in French is used to mark elision, its usage will sometimes be misinterpreted by grammarians in *a posteriori* explanations, as in the case of the adjective *grand'* (tall) "[the] adjective *grand* was originally an adjective of one termination (*grandem* in Latin giving in French to one form for both masc. and fem.) When later grand by analogy with other adjectives, assumed the feminine e, the grammarians thought they saw in the absence of the e in the older form a mark of elision, which they indicated by an apostrophe, hence such forms as: *grand' mère, grand' route, grand' messe, grand' peur, grand' peine, grand' chose,…*" (Crane 1907, 269).

**Great Britain:** The apostrophe appears in the first original English cosmography of the 16th century, *The Cosmographical Glasse* written by the physician and cartographer William Cunningham and published in 1559.

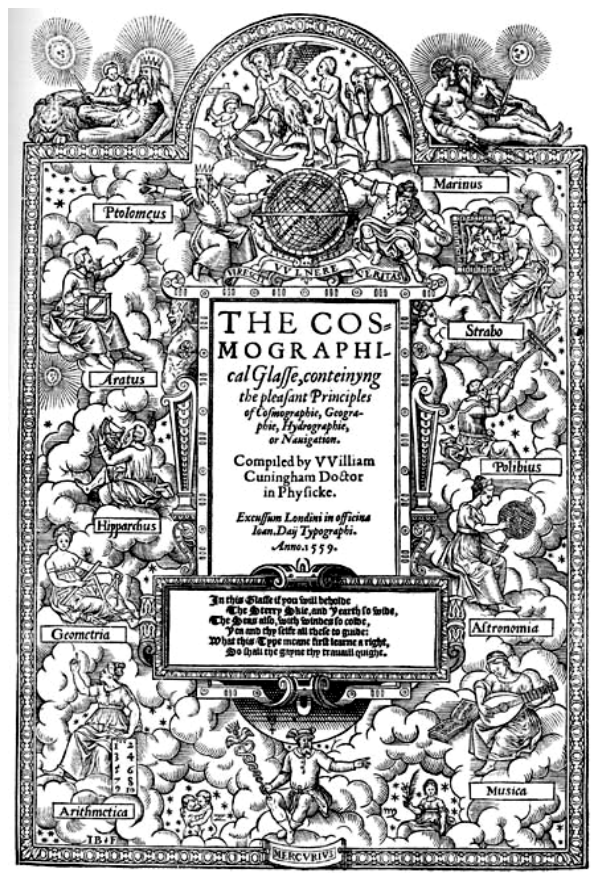

Illustration 6: Frontispiece of *The Cosmographical Glasse*

Illustration 7 shows a common typographical abbreviation (ō for on) and the final e in the definite article "the" in "the earth" is elided. But generally speaking how is the apostrophe defined in the 17th century?

---

[3] As De Vinne (1886) explains "he traced the derivation of all forms of the alphabetical letters to the two letters which make the name of the mythological goddess IO. From this straight line and circle came all the letters. He made the human figure fit into a geometrical diagram on which he planned the shapes of letters" (34).

[4] « Ie dis et allegue ces choses icy afin que sil avenoit quon deust escripre en lettre attique telz metres ou le S se debvroit evanoyr, on les porroit escripre honnestement et scientement sans y mettre ladicte lettre…, et escripre ung point crochu au-dessus du lieu ou elle debvroit estre » ("I'm saying and asserting these things so that, if it so happened that one needed to write in Attic letters some verse where the S should be elided, one could omit the aforesaid letter and instead —according to correct and scholarly usage— write a hooked dot above the place where it should have appeared", G. Tory, *Champ fleury* (Paris, 1529) fol. 56 v°. As the reader can see, Tory himself does not use the apostrophe much in this text, but mentions its use in Latin.

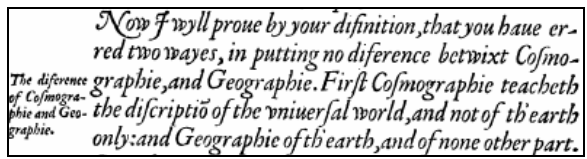

Illustration 7: Excerpt from *The Cosmographical Glasse*

A *Dictionarie of the French and English Tongves*, compiled by Randle Cotgrave in 1611, the first French-English bilingual dictionary, includes the following entry: "To apostrophise is to cut off the last vowel of a word" (cf. Illustration 8). Actually we shall see that not only vowels could be "cut off" or elided, but the apostrophe could have another function that will be examined in part two of this essay.

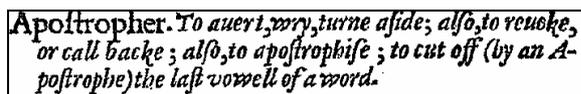

Illustration 8: Definition of *Apostropher* in Cotgrave's *Dictionary of the French and English Tongves* (1611)

### 2.2 General Remarks about its present use

**Comparison between European languages**: If we compare the uses of the apostrophe and their occurrences in European languages today, André (2008) says that whatever big corpuses in different European languages we use —for instance sacred or lay texts such as the Gospels or the project of the European Constitution— we always get similar results, and in the following proportions:

| Language | French | Italian | English | German | Spanish |
|---|---|---|---|---|---|
| Number of apostrophes | 10 000 | 6 000 | 500 | Some | Some |
| Number of sentences | 9 000 | 9 000 | 9 000 | 9 000 | 9 000 |

Table 1: apostrophes in European languages, a comparative chart

**Encoding:** To conclude this brief historical outline, we ought to mention a difficulty which brings us back to technological issues. A problem occurs when apostrophes are processed by computers. The apostrophe looks very much like the single quotation mark, and it is typed with the same key. Different computer encoding systems have different encodings for the apostrophe. Here are the codes that can be found on the Internet: 39 (not used by Anglo-Saxons); 169; Unicode U 2018 or U 2019. Codes 2018 and 2019 cannot be both typed with the keyboard when we use NooJ. So in some texts one might lose a few results when one searches a text using one specific code, whereas different codes might be employed in the text. Maybe that is why search engines remove or ignore punctuation marks when they process your request.

Nevertheless, as we shall see below, it is possible to create efficient NooJ tools (in the form of graphs or dictionaries) to process 17th century texts.

### 3 Uses and functions of the apostrophe

Our corpus includes a selection of English travel accounts in Greece and Anatolia in the 17th century. For accredited researchers these texts are available for download from the EEBO (Early English Books Online) database. They have been scanned as images and saved in pdf format. Therefore it was necessary to type them before they could be computed by NooJ.

The authors we have elected to study are (in chronological order) George Sandys, Henry Blount, John Ray, Paul Rycaut, Thomas Smith and George Wheler. Their narratives, which were "best-sellers" in their day, provide us with an interesting sample of 17th century English travel literature.

How do apostrophes feature in our corpus? We will start by spotting the occurrences of apostrophes in our texts. We shall see that they had specific uses and functions that sometimes differ from modern usage. Then we will analyse and transcribe them thanks to the morphological and syntactical graphs we have created with NooJ. In the meantime, we will be discussing our methodology and the technical problems we are still confronted with.

As in contemporary English, the apostrophe may signal an abbreviation in conjunctions and prepositions or the elision of a letter in nouns, adverbs[5] and verbs, but it is much more frequently used.

---

[5] Ever and even may be spelled e'er and ev'n; we haven't found any instances of these forms in our corpus.

### 3.1 Conjunctions and Prepositions

Conjunctions and prepositions are often abbreviated. One or more letters may be removed at the beginning (apheresis) or at the end (apocope) of a word as in the following examples: tho' for though, ("tho' dismist the seraglio"); thro' for through ("thrusting an iron stake thro' the body out under the neck"). The grammarian Miège (1688) also mentions the following aphereses: 'bove for above, 'midst for amidst and 'twixt for betwixt (110).

### 3.2 Nouns

The apostrophe begins to be used as the marker of the Saxon genitive (as in "the Grand Signior's women"). It is important to remark that the genitive was previously formed without using an apostrophe and was still very common in 17[th] century texts. The following examples are noteworthy:

"from the *womens* apartment";
"out of their *wives and childrens* mouths".

The apostrophe was also used to form the interlingual plural of foreign nouns in words ending in a vowel:[6]

- *Egoumeno'*s, *capricio*'s
- *Bassa'*s (Pashas),
- *piazza*'s

These lemmas can be spotted thanks to a NooJ grammar, and thanks to the graph below:

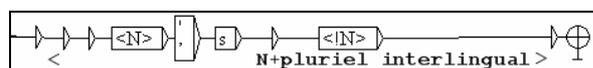

Graph 1: identifying the interlingual plural

**Unknown tokens:** But a problem occurs for unknown words ending in s, as in the following examples:
-*anothers:* "one anothers company"
-*childrens:* "their wives and *childrens* mouths"

Therefore we created a morphological graph —graph 2— to allow for the recognition of these forms:

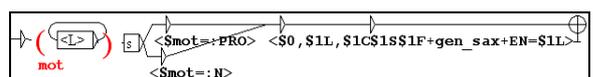

Graph 2: identifying the Saxon genitive

It so happens that the form "wives" can also be a genitive. In this case, the conjunction "and" will help us to recognise it thanks to the syntactical graph that transcribes it. Graph 3 processes the Saxon genitive of words identified by graph 2, and it may extend transcription to a word linked by a conjunction and to a word recognised by graph 2.

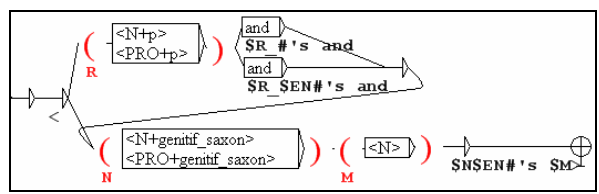

Graph 3: transcribing the Saxon genitive

In a nutshell, this graph can recognise:
- either noun + noun : <N+gen_sax> <N>
- or pronoun + noun: <PRO+gen_sax> <N>
- wives and childrens mouths → < wives's[7] and children's mouths>
- anothers company → <another's company>.

But we cannot automatise other cases.

**Case of "mans":** So much for unknown words, but what about known words? Can they be recognised and correctly interpreted?

Let's take the example of "mans" in the following sentence: "*First, that no other mans errors could draw either hatred, or engagement upon me*". "Mans" is not identified as a genitive since it is wrongly recognised as the transitive verb "to man". There is a solution to this problem, by including this entry (to man) in our dictionary, as the list of uses of the verb to man is rather short (to man a ship; a fort; the guns).

### 3.3 Verbs

The apostrophe is used in the contracted forms of verbs as in contemporary English. For the verb to be, 'tis (no longer in use in CE) is contracted into "it is" or it's. To recognise it with NooJ we need to annotate the word form as a sequence of annotations:

'tis → <it,PRO+3+s> <is,be,V+PR+3+s>

Miège (see annex, illustration 9) mentions other abbreviations after the same model such as 'twere, 'twill, 'twould, 'twas, ben't, 'ent and several other abbreviations for I will, I would and I had.

---

[6] According to Miège 's was also used to form the plural of English words ending in y, hence the plural of heresy might be heresy's or heresies. We have not found any instance of this in our corpus.

[7] To be able to automatise the transformation by adding 's, we need to use this spelling (wives's instead of wives') which prescriptive grammarians consider incorrect.

**The "ed" flexion:** The "e" in the "ed" flexion can be elided in the preterit and past participle of all verbs, hence the mute "e" is turned into an apostrophe and omitted as in "belov'd". The final d is also spelt "t" to mirror the pronunciation of the word. The following quotation provides good examples.

> Let ravisht poets drink thrice three,
> Of whom the uneven muses be
> Belov'd. The grace misdoubting jarres,
> Linkt to her naked sisters, barres
> Draughts that exceed that number.
> *Horat. 1.3 od 19.*

Whence we notice that the apostrophe also indicates the elision of "e" in the preterit and present perfect of verbs ending in p, k, x, ss, sh, which means that these verbs have four forms, one in ed and shortened forms in 'd, 't or t. For regular forms and for shortened forms in t or d, we use a morphological graph and a syntactical graph applied consecutively. Different cases crop up. Known words and unknown words require different transformations.

The following morphological graph (4) may recognise an unknown part of a verb by adding ed, ted or sed:

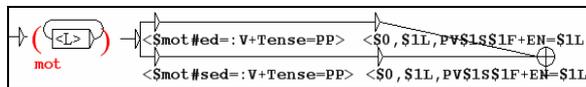

Graph 4: recognising elided forms of verbs

Then the following syntactical graph is applied:

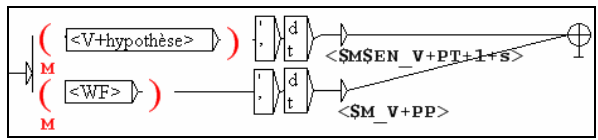

Graph 5: transcribing elided forms of verbs

Thus a verb like establish can have 4 forms:
"and *establish'd* the same number",
"the *establish't* doctrine",
"and for ever *established* the adoration".
"*establisht*"

The dictionary first recognises **establish** as the bare infinitive or the simple present of the verb to establish. So the syntactical graph must modify the result of the dictionary. Then the syntactical graph builds the form "established".

What about verbs which are not recognised or wrongly recognised? Let's examine different cases. These three cases have been selected because they correspond to three different situations: 1. An unknown token (judg); 2. A token (profes) wrongly identified as a noun when performing a basic transformation, i.e. eliding a mute e.; 3. An unrecognized token (imbrac) requiring a morphological transformation.

**Unknown token judg:** The form "judg" in "this being judg'd" is not recognised. So the morphological graph tests some hypotheses:
"judg" + ed → judged = the PP of judge
At that point we don't know whether judg is followed by 'd in the text. Graph 5 will confirm it and we get the following result:
judg,judge,V+Tense=PP+ EN=judge

**Wrong identification of tokens:** Some tokens may be falsely recognised as nouns as in the following example: "Though the Christian religion be profes'd in the Ottoman dominions".
The token "profes" is unknown. A first graph makes a simple change that consists in deleting a mute e in profes and suggests it is the plural of the noun prof (as the NooJ dictionary includes this entry):
profes,prof,N+Nb=p+Distribution=Hum
Then another graph, graph 4, adds /sed/ and suggests "professed" (the full graph also "recognises" a preterit and an adjective):
profes,profess,PV+Tense=PP+EN=profess
Then graph 5 invalidates the noun, and validates the verb:
profes,profess,V+Tense=PP+EN=profess

**Case of imbrac'd or imbrac't:** To be able to recognise imbrac'd NooJ needs to combine two transformations:
imbrac'd → embrac'd → embraced.
The last change is made by graph 5, but the first one needs to explore the ninety transformation graphs that we have drawn, to be able to recognise this form. All the graphs will produce several answers from which the correct one must be chosen. This means our results need to be validated and the correct answer chosen.

**Hierarchy of graphs:** Several morphological graphs with different levels of priority may be applied to unknown tokens. The hierarchical organisation of our graphs means that once a solution has been found, the other graphs with less priority will not be applied. The option that con-sists in applying all the graphs at the same time has the undesirable effect of producing myriads of wrong answers. Getting the priorities right when using syntactical graphs is a difficult problem that we have not sorted out yet.

### 3.4 Apostrophes in our corpus: charts

Our corpus has 91 occurrences of 'd or 't.

| Author | Year | Size of corpus | Different words | Number of ' |
|---|---|---|---|---|
| Blount | 1636 | 1408 | 621 | 0 |
| Sandys | 1652 | 3910 | 1419 | 15 |
| Rycaut | 1679 | 12920 | 2270 | 11 |
| Smith | 1682 | 22640 | 4285 | 113 |
| Wheler | 1682 | 2365 | 747 | 4 |
| Ray | 1693 | 1140 | 533 | 15 |
| **Total** | | **44383** | | **158** |

Table 2: Global information about our corpus

| Author | Number of 's | genitive | plural |
|---|---|---|---|
| Sandys | 3 | 3 | 0 |
| Rycaut | 8 | 8 | 0 |
| Smith | 40 | 27 | 13 |
| Wheler | 2 | 1 | 1 |
| Ray | 1 | 0 | 1 |
| **Total** | **54** | **39** | **15** |

Table 3: Occurrences of 's

| Author | Number of 'd | Other ' |
|---|---|---|
| Sandys | 11 (in poems) | 1 th' (poem) |
| Rycaut | 3 | 0 |
| Smith | 63 + 1 't | 4 tho' 1 thro' 4 'tis |
| Wheler | 2 | 0 |
| Ray | 11 | 1 tho' 2 'tis |
| **Total** | **91** | **13** |

Table 4: Occurrences of 'd and other cases

There are no apostrophes in the earliest text, but the excerpt is very short. We cannot infer from our corpus general conclusions about the frequency of its usage, according to chronology or date of publication. Usage depends on the author's style. In other 17[th] century texts such as Milton's pamphlets (published in the 1640s) or Mary Astell's essays (published in the 1690s), usage of the apostrophe is widespread, and usually marks contractions in verbs, the elision of many unpronounced letters (vowels or consonants) or mute e's but also the plural of certain nouns.

## 4 Conclusion

In natural language processing, transformations involving multiple tokens create more difficulties than those who are merely the transformation of one token (such as the introduction of a mute e or doubling of a letter). The problem to address here is technically very different if the modernisation of the text requires adding an apostrophe, or deleting it. Processing it involves two steps, which makes it more difficult. Processing 't or 'd means dealing with a verb or an adjective sometimes, whereas 's may appear only in a noun. If it is an interlingual plural, we just delete the apostrophe and concatenate the s.

In cases where one must insert 's —i.e. in nouns— a problem occurs when recognising the word, if the latter can be a noun or a verb. How can we recognise the genitive? We need to insert the apostrophe: mans→man's. In the case of prepositions, pronouns and conjunctions which are after all few in number, we can process them simply by adding them to our dictionary when they are not ambiguous.

The introduction of the apostrophe into a verb splits it into two tokens, and alas the fact that the apostrophe is a delimiter makes our work very complicated. But this is strictly a computer problem that automatic processing does not allow us to solve immediately. The beginning of the truncated word will be falsely recognized, and that will produce an erroneous annotation, which must be removed afterwards, or it will be labelled UNKNOWN. Then the application of syntactic grammars will allow for the identification and removal of the apostrophe, provided that the verb is in the NooJ dictionary (we may complete it if necessary). Thus two NooJ tools are used here: some forms are recognised by simply inserting the short forms in the dictionary, while for nouns and verbs the use of morphological and syntactic graphs is necessary.

The morphological and syntactical graphs we have drawn may process all of the occurrences of the apostrophe in our corpus, with very little noise. Usage of the apostrophe seemed simpler in 17[th] century English. The possessive genitive in 's was not very common yet. The apostrophe mostly marks the omission of letters in a wide range of words and the plural of certain words.

French students often find 17[th] century English texts difficult to read and understand. We hope our tools (especially the dictionaries) might make these texts more accessible to students of English as a second language.

# ANNEX

| | | |
|---|---|---|
| he's | | he is: |
| in't | | in it. |
| 'tis | | it is. |
| 'twas | | it was. |
| 'twere | | it were. |
| 'twill | | it will. |
| 'twould | | it would. |
| th' | | the. |
| ith' | for | in the. |
| oth' | | on the. |
| t'other | | the other. |
| 'em | | them. |
| e're | | ever. |
| ne're | | never. |
| o're | | over. |
| e'en | | even. |
| on't | | of it. |
| don't | | do not. |
| han't | | have not. |
| shan't | | shall not. |
| can't | | cannot. |
| ben't | for | be not. |
| 'ent | | is not. |
| d' ye | | do ye. |
| I'l, I'le, I'll | | I will. |
| I'd, I'de | | I would, or I had. |

Illustration 9: A list of contractions
(Miège 1688, 110-11)